\documentclass[letterpaper,10pt,conference]{ieeeconf}

\usepackage{iftex}
\ifXeTeX
  \usepackage[OT1]{fontenc}
\fi

\IEEEoverridecommandlockouts
\usepackage{amsmath}
\usepackage{amssymb}
\usepackage{booktabs}
\usepackage[font=footnotesize]{caption}
\usepackage{subcaption}
\usepackage{cuted}
\usepackage{dblfloatfix}
\usepackage{afterpage}
\usepackage{graphicx}
\usepackage{hyperref}
\usepackage{tabularx}
\usepackage{xcolor}
\usepackage{xspace}

\definecolor{modelOurs}{HTML}{D65E83}
\definecolor{modelPi}{HTML}{DFAB72}
\definecolor{modelRDT}{HTML}{9F95BF}
\definecolor{modelFastWAM}{HTML}{6AAEBC}
\definecolor{modelLingBot}{HTML}{94B78F}
\newif\ifexperimentcolors
\experimentcolorsfalse
\newcommand{\modelname}[2]{\ifexperimentcolors\textcolor{#1}{\textbf{\boldmath#2}}\else#2\fi}
\newcommand{\ours}{\modelname{modelOurs}{\emph{Spatial-UMI-Dream}}\xspace}
\newcommand{\pifive}{\modelname{modelPi}{$\pi_{0.5}$}\xspace}
\newcommand{\rdtfm}{\modelname{modelRDT}{RDT2-FM}\xspace}
\newcommand{\rdt}{\modelname{modelRDT}{RDT2}\xspace}
\newcommand{\fastwam}{\modelname{modelFastWAM}{Fast-WAM}\xspace}
\newcommand{\lingbot}{\modelname{modelLingBot}{LingBot-VA}\xspace}
\newcommand{\simname}{\textit{RoboTwin-UMI}\xspace}
\newcommand{\totalTime}{605}

\newcommand{\figref}[1]{Fig.~\ref{#1}}
\newcommand{\tabref}[1]{Table~\ref{#1}}

\title{\LARGE \bf Spatial-UMI-Dream: Scaling World Action Models with Metric Geometry for UMI Manipulation}
\author{
   Weiduo Yuan$^{1,*}$\quad
   Ze Yuan$^{1,*}$\quad
   Sihan Liu$^{2, *}$\quad
   Jionghao Wang$^{3}$\quad
   Liang Heng$^{1}$\quad
   Zhenyu Zhao$^{1}$\quad
   Yue Wang$^{1}$\\
   $^{1}$USC Physical Superintelligence Lab\quad $^{2}$CMU\quad $^3$Texas A\&M University\\
   $^{*}$Equal contribution
}
\begin{document}

\nocite{BSTcontrol:authorlists}

\maketitle
\thispagestyle{empty}
\pagestyle{empty}

\begin{strip}
  \vspace*{-12mm}
  \centering
  \includegraphics[width=\textwidth]{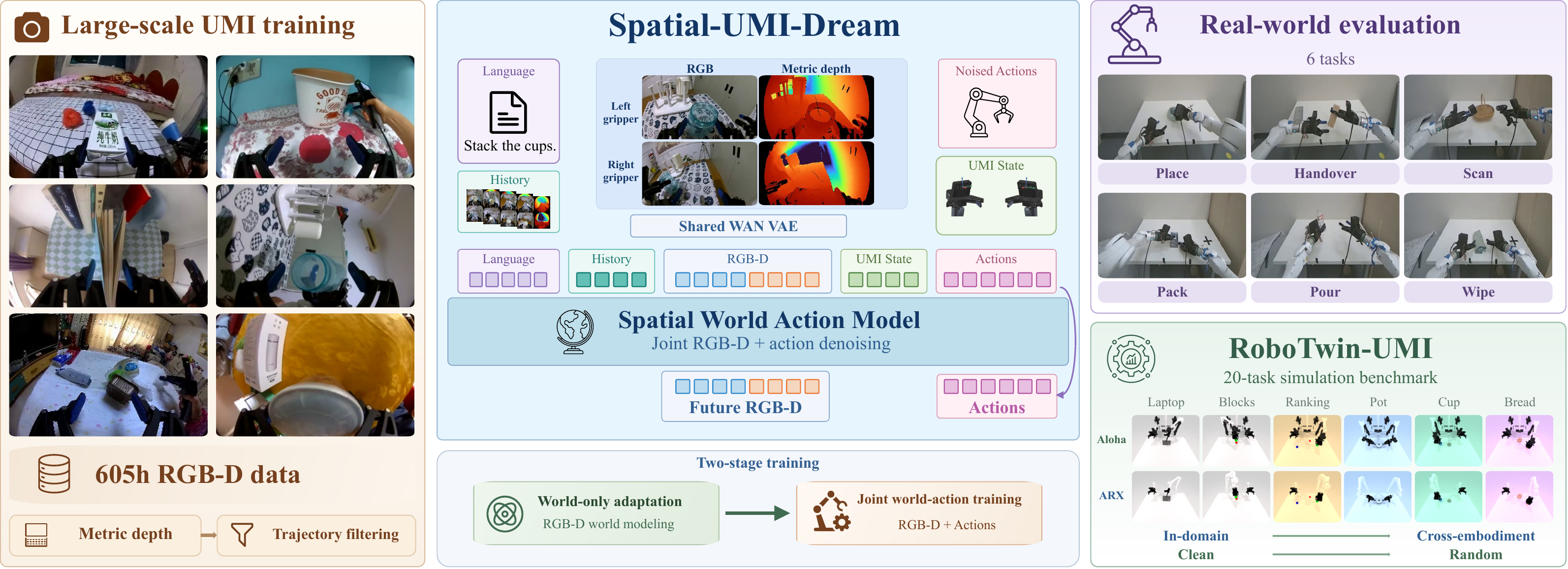}
  \captionof{figure}{\textbf{Spatial understanding and cross-embodiment transfer for UMI policies.} We introduce \ours, a 5B world-action model that learns jointly from RGB, metric depth, and actions through \totalTime{} hours of UMI pretraining. Across six real-world bimanual tasks and the twenty-task \simname benchmark, it achieves the best in-domain performance and generalizes best across robot embodiments and environments.}
  \label{fig:teaser}
\end{strip}

\begin{abstract}
The Universal Manipulation Interface (UMI) enables scalable demonstration collection independently of a specific robot.
However, its moving, close-range gripper cameras entangle camera and object motion without a stable external view, leaving the metric geometry needed for precise manipulation implicit in RGB.
We introduce \ours, a 5B spatial World Action Model (WAM) pretrained on \totalTime{} hours of RGB-D UMI data through two-stage training---world-only adaptation followed by joint world-action learning---to jointly predict future RGB, fisheye-aligned metric depth, and actions.
Stereo-derived depth makes scene scale and spatial relationships explicit within a unified dual-gripper RGB-D representation.
Furthermore, evaluating UMI policies requires testing whether UMI's robot-independent demonstration collection translates into policy generalization across robot embodiments, alongside robustness to environmental changes.
We introduce \simname, a twenty-task simulation benchmark built on RoboTwin with UMI wrist-mounted fisheye observations, enabling controlled, repeatable evaluation of UMI policies under environment and embodiment changes separately and jointly without additional fine-tuning.
On \simname, \ours achieves 79.10\% in-domain success and 72.15\% under combined embodiment and environment shifts, exceeding the strongest evaluated baseline in each setting by 4.65 and 10.30 percentage points, respectively.
It also exhibits the smallest success-rate drop under embodiment transfer among the evaluated policies.
In a matched pretraining ablation, RGB-only pretraining leads to a 5.75-percentage-point decrease in in-domain success compared with RGB-D pretraining.
Across six real-world tasks, \ours achieves 85.0\% average success, exceeding the strongest baseline by 24.2 percentage points.
Our project page is available at \url{https://weiduoyuan.com/spatial-umi-dream/}.
\end{abstract}

\section{INTRODUCTION}
\label{sec:introduction}

The Universal Manipulation Interface (UMI)~\cite{chi2024universalmanipulationinterfaceinthewild} enables scalable collection of manipulation demonstrations without requiring a robot at the demonstration site. Its portable handheld grippers record close-range video and embodiment-agnostic end-effector trajectories during in-the-wild human manipulation. These demonstrations provide interaction data for learning policies that can execute manipulation skills across tasks, environments, and robot embodiments.

Vision-Language-Action models (VLAs) and World Action Models (WAMs) offer two approaches to learning manipulation policies. VLAs~\cite{intelligence2025pi05visionlanguageactionmodelopenworld,wei2026psi0,liu2026rdt2exploringscalinglimit,kim24openvla} build on pretrained vision-language models to predict robot actions from visual observations and language instructions. WAMs~\cite{ye2026worldactionmodelszeroshot,nvidia2026cosmos3omnimodalworld,dyna2026dyna2} instead build on pretrained video generation models to jointly predict future visual observations and actions. This joint prediction connects actions with the resulting scene changes. Future video supplies supervision across space and time, while pretrained video diffusion models provide priors over motion and scene dynamics.

Although WAMs have shown promising results with RGB observations, applying them to UMI data presents two challenges. First, RGB-only models leave the metric geometry needed for manipulation implicit in appearance and image-space motion. UMI has no fixed third-person view, and its gripper-mounted fisheye cameras move with the grippers during close-range manipulation. Changes in the images therefore reflect camera motion, object motion, and scene geometry. In bimanual manipulation, both views move independently, while the model must coordinate two grippers in a shared three-dimensional workspace. The model must therefore infer scene scale and spatial relationships from RGB alone. Second, in-domain task success alone is insufficient to evaluate UMI policies. Lighting and scene appearance can vary substantially in wrist-camera observations, and demonstrations collected in a limited set of environments may not cover these variations. Moreover, because UMI decouples demonstration collection from a specific robot, cross-embodiment transfer is central to evaluating UMI policies. Evaluation must therefore test both robustness to environmental changes and the ability to transfer learned skills across robot embodiments.

To address these challenges, we introduce \textbf{\ours}, a spatial WAM that incorporates metric depth, and \textbf{\simname}, a benchmark for evaluating UMI policies across environments and robot embodiments.

\textbf{\ours} is a 5B spatial WAM pretrained on \totalTime{} hours of RGB-D UMI data to jointly predict future RGB, metric depth, and robot actions. Training proceeds in two stages: world-only adaptation to UMI RGB-D observations, followed by joint world-action learning. We recover metric depth from calibrated stereo observations and align it with the raw fisheye RGB images. Metric depth makes scene scale and the spatial layout around each gripper explicit, reducing the need to infer this information from RGB alone. We arrange both grippers' aligned RGB-D observations in a unified video canvas, incorporating geometry into world-action modeling while retaining the pretrained video tokenizer and WAM backbone.

\textbf{\simname} is a twenty-task benchmark built on RoboTwin~\cite{chen2025robotwin} that reproduces UMI's wrist-mounted fisheye observations. Simulation enables controlled, repeatable evaluation of the same tasks across different environments and robot embodiments. This supports testing both environmental robustness and cross-embodiment transfer, which is particularly important because UMI demonstrations are collected independently of a specific robot. The benchmark provides a common closed-loop protocol for comparing UMI policies in both in-domain and generalization settings. Environment and embodiment shifts are tested separately and jointly under a consistent UMI observation interface, with all generalization tests using the trained policies without additional fine-tuning.

\ours achieves 79.10\% in-domain success on \simname and 85.0\% average success across six real-world tasks, exceeding the strongest evaluated baselines by 4.65 and 24.2 percentage points, respectively. Its advantage extends to generalization: the lead over the strongest baseline widens to 14.00 percentage points after embodiment transfer, and \ours remains the strongest policy under environmental and joint shifts.

Our contributions are twofold:
\begin{itemize}
    \item \textbf{Spatial world-action modeling at scale.} We introduce \ours with two-stage training on \totalTime{} hours of RGB-D UMI data, achieving the highest in-domain and generalization success rates among evaluated policies. Ablations demonstrate the benefit of metric depth and progressive gains from the two training stages.
    \item \textbf{A benchmark for UMI policy generalization.} We develop the twenty-task \simname benchmark with wrist-mounted fisheye observations and a common protocol for evaluating in-domain performance and generalization across environments and robot embodiments, including their combined shifts.
\end{itemize}

\begin{figure*}[t]
  \centering
  \includegraphics[width=0.98\textwidth,trim=0 0 0 -16bp]{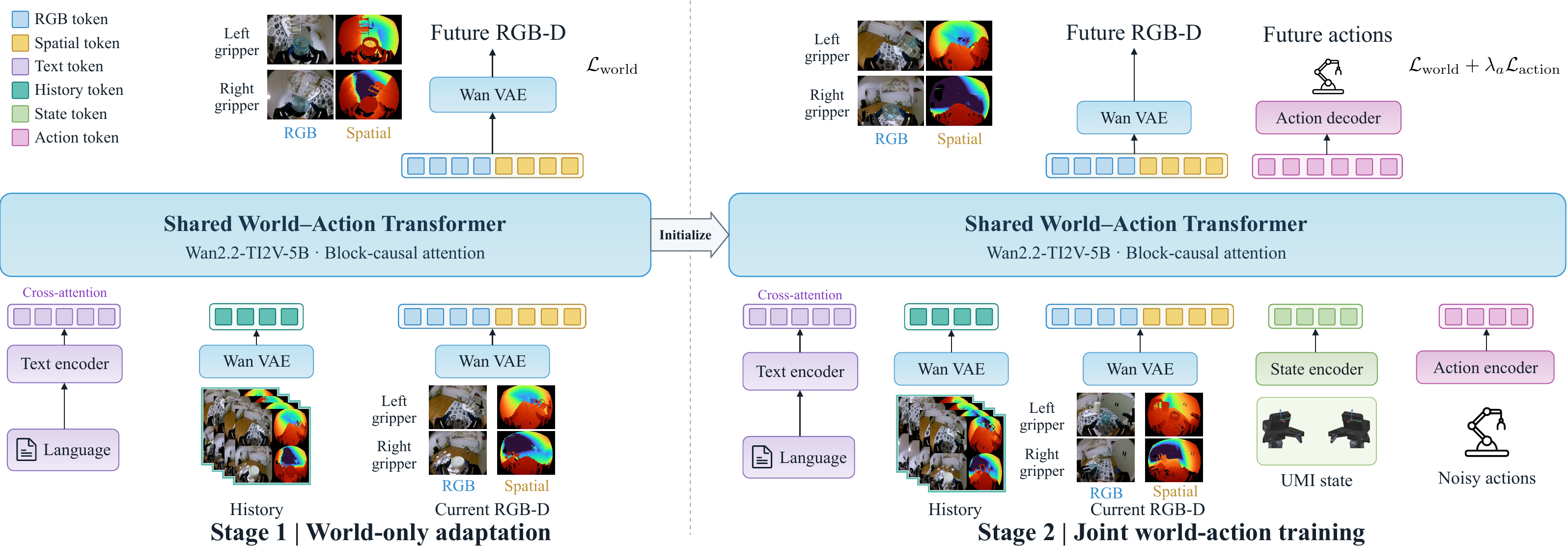}
  \caption{\textbf{Model architecture and two-stage training of \ours.} In both stages, dual-gripper RGB and metric depth (Fig.~\ref{fig:umi-data}) form a unified RGB-D canvas. The Wan VAE encodes observation history and current RGB-D inputs into visual tokens, while a text encoder supplies language tokens through cross-attention. The shared world-action transformer uses \textbf{block-causal attention} to condition prediction on preceding visual history. \textbf{Stage 1:} world-only adaptation predicts future RGB-D observations, decoded through the Wan VAE and supervised by $\mathcal{L}_{\mathrm{world}}$. \textbf{Stage 2:} the transformer backbone is initialized from the Stage 1 world-only adaptation checkpoint. Joint world-action training additionally encodes UMI state and noisy actions, and jointly predicts future RGB-D observations and actions through the Wan VAE and action decoder, optimized with $\mathcal{L}_{\mathrm{world}}+\lambda_a\mathcal{L}_{\mathrm{action}}$.}
  \label{fig:method-overview}
\end{figure*}

\section{RELATED WORK}
\label{sec:related-work}

\subsection{Robot Foundation Models}
\label{subsec:related-foundation-models}
Robot foundation models seek to reuse broad visual-language knowledge for control~\cite{kim24openvla,wu2026largerewardmodelsgeneralizable}, enabling policies to interpret diverse scenes and instructions without learning each task in isolation.
Representative action-centric Vision-Language-Action models~\cite{intelligence2025pi05visionlanguageactionmodelopenworld,wei2026psi0,liu2026rdt2exploringscalinglimit,kim24openvla} adapt pretrained vision-language representations to predict robot actions from visual observations and language instructions.
However, predicting actions alone does not explicitly describe how the scene is expected to evolve, leaving future outcomes unavailable as a direct policy context or planning signal.
More recent methods~\cite{intelligence2026pi07steerablegeneralistrobotic,jia2026dreamplanefficientreinforcementfinetuning,zhang2026foreactsteeringvlaefficient} introduce world models into this paradigm without fully coupling world and action generation.
By using predicted visual futures as policy context or planning supervision, these approaches improve policy generalization and planning while retaining a separation between world prediction and action generation.
World Action Models~\cite{ye2026worldactionmodelszeroshot,nvidia2026cosmos3omnimodalworld,dyna2026dyna2,bi2025motusunifiedlatentaction} instead jointly generate future observations and actions, directly connecting predicted scene evolution with control.
Traditional RGB-only WAMs leave metric geometry implicit in visual appearance. In contrast, \ours jointly predicts RGB, registered metric depth, and actions, making metric geometry an explicit component of the world state.

\subsection{Geometry-Aware Robot Learning}
\label{subsec:related-geometry}
Geometric representations support spatial reasoning in robotic perception and control~\cite{bevcalib,Ze2024DP3,3d_diffuser_actor,nguyen2026iclrincontextimitationlearning}, providing explicit cues about scene layout and spatial relationships.
3D visuomotor policies explicitly encode depth or point clouds into geometric scene features, allowing action prediction to operate on metric structure rather than infer it only from visual appearance.
Spatially grounded Vision-Language-Action models~\cite{zhen20243dvla,qu2025spatialvla} further incorporate explicit spatial representations into language-conditioned action prediction.
Representation-aligned policies~\cite{li2026spatialforcing,sun2026rocket} instead distill geometric features from pretrained 3D models into RGB policy backbones, providing spatial supervision without requiring explicit 3D inputs at deployment.
In contrast, \ours treats metric depth as an observed and predicted component of the world state rather than only as a policy input or representation-level training target.
It jointly predicts dual-gripper RGB, metric depth, and actions through a unified video canvas, without a separate 3D encoder or alignment teacher.

\subsection{UMI Data and Evaluation}
\label{subsec:related-benchmarks}
Diverse robot datasets and standardized evaluation platforms~\cite{zhao2026humanoideverydaycomprehensiverobotic,wei2026simple} support systematic comparison of learned manipulation policies.
UMI~\cite{chi2024universalmanipulationinterfaceinthewild} enables robot-independent demonstration collection by recording wrist-mounted fisheye observations and end-effector trajectories with a portable handheld gripper.
Large-scale robot policies~\cite{liu2026rdt2exploringscalinglimit,generalist2026gen15,zhang2026hy} leverage UMI and UMI-style handheld demonstrations for generalist control and zero-shot cross-embodiment transfer.
More recent UMI-oriented VLA training~\cite{yang2026vista} studies fisheye-aware visual grounding and embodiment-conditioned physical validation to reconcile handheld demonstrations with pretrained visual representations and target-robot constraints.
UMI demonstrations collected across diverse scenes support learning policies beyond a single environment, motivating evaluation under environmental changes.
Moreover, because UMI decouples demonstration collection from a specific robot, cross-embodiment evaluation tests whether this flexibility also carries over to the learned policies.
We introduce \simname, a benchmark built on RoboTwin~\cite{chen2025robotwin} that evaluates these two capabilities separately and jointly through a controlled, multi-task simulation protocol under a consistent wrist-mounted fisheye observation interface.
This enables repeatable comparisons of different policies on the same tasks.

\section{METHOD}
\label{sec:method}

We study language-conditioned manipulation from dual-gripper UMI observations. At timestep $t$, the observation is $o_t=\{I_t^e,D_t^e\}_{e\in\{0,1\}}$, where $I_t^e$ and $D_t^e$ are the RGB image and aligned metric depth in gripper $e$'s native fisheye view. Given visual history, robot state $s_t$, and language instruction $\ell$, our model jointly predicts future RGB-D observations and temporally aligned robot action chunks for coordinated manipulation with both grippers.

Our method constructs a metric RGB-D representation, jointly models visual observations and actions, and learns the policy through two-stage training. Trajectory filtering and deployment command compensation connect handheld demonstrations to robot execution.

\subsection{Metric RGB-D Representation}
\label{subsec:spatial-wam}
\label{subsec:metric-rgbd}

\subsubsection{Metric depth reconstruction}
\begin{figure}[t]
  \centering
  \includegraphics[width=\linewidth,trim=0 0 0 -22bp]{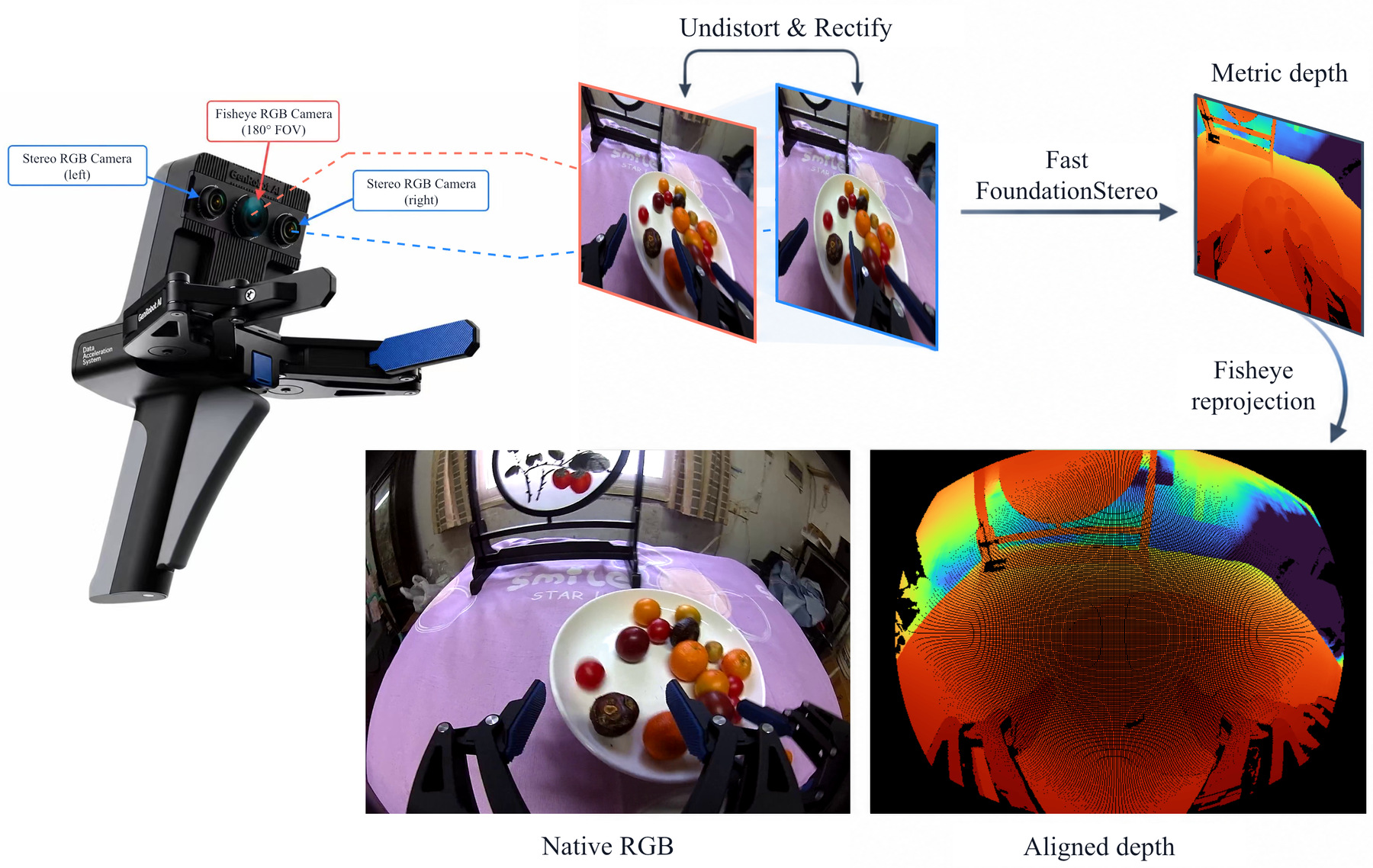}
  \caption{\textbf{Metric depth construction for UMI observations.} Each UMI gripper has three cameras. We undistort and rectify the middle and right camera views, then use Fast FoundationStereo to recover metric depth. We map the depth back to the original middle-camera fisheye view to align it with the RGB observation.}
  \label{fig:umi-data}
\end{figure}

We use the GenRobot DAS UMI system, with three calibrated cameras mounted on each gripper. As illustrated in Fig.~\ref{fig:umi-data}, the middle and right cameras form a stereo pair, with the middle camera providing the reference RGB observation $I_t^e$. We epipolar-rectify the two images and use Fast FoundationStereo~\cite{wen2026fastfoundationstereo} to estimate disparity in the rectified middle-camera image coordinates. We convert disparity to metric $Z$-depth using the stereo calibration, then map the depth back to the middle camera's original image coordinates, accounting for its lens distortion. The resulting $D_t^e$ is aligned with the native fisheye RGB observation used by the policy.

\subsubsection{Three-channel depth encoding}
Inspired by Marigold~\cite{ke2023repurposing}, we encode depth through the pretrained image VAE used for RGB. We normalize metric depth using fixed bounds $d_{\min}$ and $d_{\max}$ shared across the dataset, then apply a three-channel color mapping $\phi$:
\begin{equation}
\Phi(D) = \phi\!\left(\frac{d_{\max}-\operatorname{clip}(D,d_{\min},d_{\max})}{d_{\max}-d_{\min}}\right).
\label{eq:depth-encoding}
\end{equation}
We choose Turbo as the color mapping $\phi$. The fixed bounds maintain a consistent mapping from physical distance to encoded values. Depth preprocessing is identical during offline data construction and online deployment.

\subsubsection{Dual-gripper RGB-D canvas}
We organize the two RGB views and their encoded depth maps into an eye-paired spatial canvas,
\begin{equation}
    C_t =
    \begin{bmatrix}
        I_t^0 & \Phi(D_t^0) \\
        I_t^1 & \Phi(D_t^1)
    \end{bmatrix}.
\end{equation}
The rows correspond to the two grippers and the columns to appearance and geometry, so each RGB observation remains spatially adjacent to the depth from the same eye. Each panel is resized to $160\times160$, producing a $320\times320$ frame with a fixed layout across training and deployment. We encode the complete canvas as an ordinary video frame, $z_t=\mathcal{E}_{\mathrm{vid}}(C_t)$, using the pretrained Wan2.2 VAE~\cite{wan2025wanopenadvancedlargescale} rather than separate modality-specific encoders. Appearance and encoded geometry consequently follow the same video codec, spatiotemporal patch embedding, and video objective, without a depth-specific branch or explicit 3D backbone.

\subsection{Joint World--Action Modeling}
\label{subsec:joint-world-action}

As shown in Fig.~\ref{fig:method-overview}, action and state encoders map their inputs into the shared transformer token space, while language provides conditioning through cross-attention. The transformer jointly denoises video and action representations, which are decoded into RGB-D canvases and executable actions by the video VAE and action decoder, respectively. With block-causal attention, generation proceeds in temporally aligned RGB-D latent blocks $Z_i$ and action chunks $A_i$:
\begin{equation}
    (Z_i,A_i)\sim p_\theta\!\left(
    Z_i,A_i\mid Z_{<i},s_i,\ell\right).
\end{equation}
Here $Z_{<i}$ denotes preceding visual blocks within a finite context window. Video and action tokens interact bidirectionally within each block and attend to preceding visual blocks, allowing visual history to guide the generation of the current world-action block.
During training, we use teacher forcing~\cite{ye2026worldactionmodelszeroshot}, conditioning each noisy world-action block on clean, ground-truth visual history.

We train the video and action streams with a weighted flow-matching objective~\cite{lipman2023flowmatchinggenerativemodeling}. For a clean target $x$ and Gaussian noise $\epsilon\sim\mathcal{N}(0,I)$, we construct $x_\tau=(1-\tau)x+\tau\epsilon$ and predict the velocity $\epsilon-x$:

\begin{equation}
    \label{eq:flow-matching}
    \mathcal{L}_{\mathrm{FM}}=\mathbb{E}\!\left[
    w(\tau)\left\|v_\theta(x_\tau,\tau;c)-(\epsilon-x)\right\|_2^2
    \right].
\end{equation}

Here $w(\tau)$ is the scheduler's noise-level weight, and $c$ denotes the conditioning inputs of the corresponding stage. Applying this objective to the RGB-D video latents and action sequences gives $\mathcal{L}_{\mathrm{world}}$ and $\mathcal{L}_{\mathrm{action}}$, respectively. In implementation, squared errors are averaged over tensor elements, and invalid action dimensions and samples without action supervision are masked.

\subsection{Two-Stage World--Action Training}
\afterpage{\begin{figure}[!t]
  \centering
  \includegraphics[width=\linewidth,trim=0 0 0 -38bp]{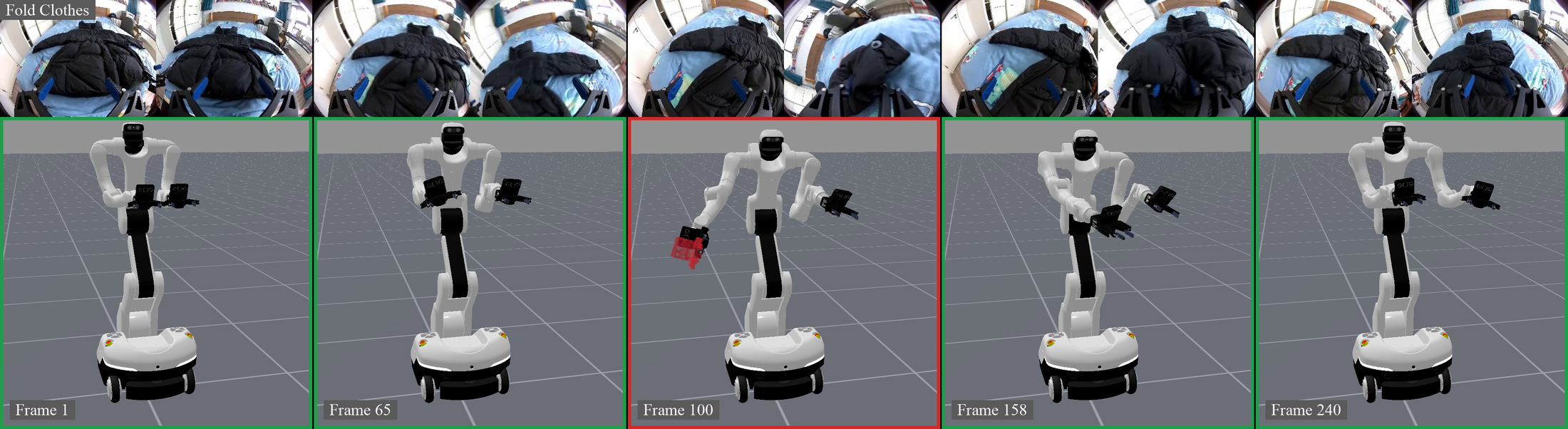}
  \caption{\textbf{UMI trajectory filtering.} Wrist observations (top) and simulation replay (bottom) of a folding demonstration. The third frame violates robot kinematic limits, so the entire episode is discarded.}
  \label{fig:trajectory-filtering}
\end{figure}
}
\label{subsec:wam-training}

We build \ours on the pretrained Wan2.2-TI2V-5B~\cite{wan2025wanopenadvancedlargescale} model and train it on the \totalTime{}-hour RGB-D UMI corpus in two stages: (1) world-only adaptation adapts the backbone to the spatial RGB-D canvas; (2) joint world-action training couples this representation with action generation. Both stages use observation history and language conditioning, with metric depth present throughout.

\subsubsection{World-only adaptation}
We optimize $\mathcal{L}_{\mathrm{world}}$ (Eq.~\eqref{eq:flow-matching}) to predict future RGB-D canvases conditioned on visual history and language.
The action branch is bypassed, leaving the action encoder and decoder untrained, and robot-state conditioning is disabled in this stage. This stage adapts the pretrained backbone to moving UMI fisheye observations and the fixed dual-gripper RGB-D layout.

\subsubsection{Joint world-action training}
We restore the world-only checkpoint and introduce the action stream and robot-state conditioning. The action encoder and decoder retain their random initial parameters, which were saved but not trained during the first stage, and begin learning jointly with the adapted backbone. The model processes noisy video and action inputs together, conditioned on observation history, language, and robot state, and minimizes

\begin{equation}
    \mathcal{L}
    = \mathcal{L}_{\mathrm{world}}
    + \lambda_a\mathcal{L}_{\mathrm{action}}.
\end{equation}

Here $\lambda_a$ controls the relative weight of action supervision.
Both $\mathcal{L}_{\mathrm{world}}$ and $\mathcal{L}_{\mathrm{action}}$ are defined by Eq.~\eqref{eq:flow-matching}.
Both stages use the same underlying RGB-D UMI corpus.

\subsection{Demonstration Filtering and Deployment}
\label{subsec:umi-pipeline}

Handheld demonstrations and robot execution differ in kinematic feasibility and gripper response. We address these differences by filtering unsuitable trajectories before training and compensating gripper commands at deployment.

\afterpage{\begin{figure}[!t]
  \centering
  \includegraphics[width=\linewidth,trim=0 0 0 -3bp]{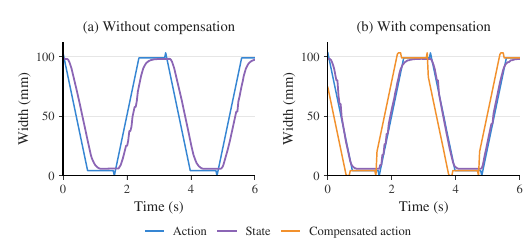}
  \caption{\textbf{Gripper compensation at deployment.} Desired actions (blue), measured states (purple), and compensated actions (orange) of one gripper (a) without and (b) with compensation under the same continuous target trajectory. Both panels show the first 6 seconds without smoothing. Mean absolute tracking error over this interval decreases from 12.49 to 2.85\,mm.}
  \label{fig:deployment}
\end{figure}
}

\subsubsection{Trajectory filtering}
Handheld UMI collection is not constrained by the target robot's kinematics or execution limits. We therefore filter demonstrations before training to avoid supervising the policy with motions the robot cannot reliably execute. We solve inverse kinematics for Dexmate Vega1 and replay the motions in ManiSkill~\cite{tao2024maniskill3} to identify and discard unsuitable episodes (Fig.~\ref{fig:trajectory-filtering}). We also discard entire episodes when any velocity, acceleration, or jerk value exceeds its dataset-wide 99th percentile.

\subsubsection{Deployment command compensation}
UMI gripper actions specify desired physical openings, but directly issuing them to the robot controller produces delayed gripper responses (Fig.~\ref{fig:deployment}(a)). To align the executed gripper state with the predicted action trajectory, following~\cite{yang2026realtimevlav2learningrun}, we calibrate a first-order response model with delay and invert it to compensate the commands:
\begin{equation}
u_k=\operatorname{clip}\!\left(\frac{r_{k+1+\delta}-\alpha r_{k+\delta}}{1-\alpha},\,0,\,w_{\max}\right),
\label{eq:gripper-compensation}
\end{equation}
Here $r_k$ and $u_k$ denote the desired width and compensated command, $\delta$ is the delay in control steps, $\alpha=\exp(-h/\tau)$ for control period $h$ and calibrated response time constant $\tau$, and $w_{\max}$ is the maximum opening. Advancing and pre-amplifying the commands aligns the physical response with the desired trajectory under the calibrated model (Fig.~\ref{fig:deployment}(b)).

\subsection{\simname Benchmark}
\label{subsec:benchmark}

\afterpage{\begin{figure}[!tb]
  \centering
  \includegraphics[width=\linewidth,trim=0 0 0 -30bp]{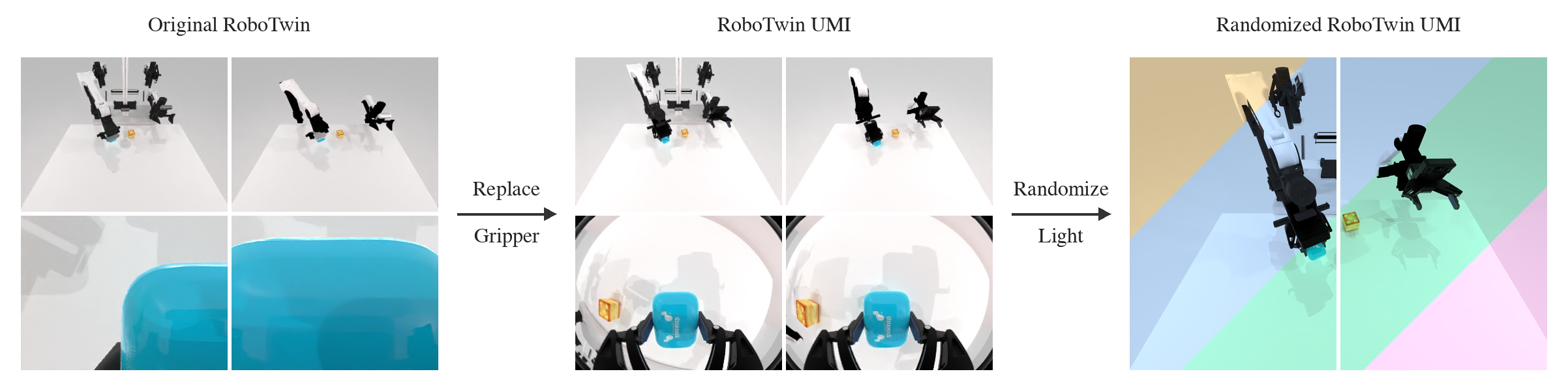}
  \caption{\textbf{\simname benchmark construction.} We retain the same RoboTwin tasks while adapting the grippers and observations for UMI. From left to right: the original setup; UMI grippers and wrist-mounted fisheye cameras on Aloha-AgileX and ARX-X5; and randomized lighting. These settings support evaluation of cross-embodiment and environmental generalization, separately and jointly.}
  \label{fig:benchmark-setup}
\end{figure}
}

To test whether the flexibility of robot-independent UMI data collection translates into cross-embodiment policy transfer, we equip two robot embodiments with a common UMI gripper and observation interface. To assess environmental robustness, we vary lighting conditions while retaining the same tasks. We introduce \simname, a twenty-task benchmark built on RoboTwin~\cite{chen2025robotwin} that evaluates these changes separately and jointly.

\subsubsection{Benchmark Construction}
We retain twenty RoboTwin tasks spanning object placement, stacking, and articulated-object interaction, and adapt the robot setups for UMI execution and observation. Specifically, we equip Aloha-AgileX and ARX-X5 with UMI grippers and wrist-mounted fisheye cameras (\figref{fig:benchmark-setup}), matching the real-world UMI camera setup. Both embodiments share the same evaluation tasks (\figref{fig:sim-results-tasks}).

\subsubsection{Fisheye Observation Rendering}
To reproduce UMI's wide-field-of-view fisheye observations in simulation, we synthesize each image from four pinhole views sharing a common optical center: one forward-facing view covers the central region, while three auxiliary views face outward in different directions to cover the periphery. For each output pixel $\mathbf p$, the calibrated Kannala--Brandt (KB4) model gives a viewing ray. We then project this ray into a pinhole image to locate the corresponding source pixel $\mathbf q_j$:
\begin{equation}
\mathbf q_j = \pi_{K_j}\!\left(R_j^\top\pi_{\mathrm{KB4}}^{-1}(\mathbf p)\right),
\label{eq:fisheye-resampling}
\end{equation}
where $R_j$ denotes camera $j$'s rotation relative to the fisheye reference frame, $R_j^\top$ maps the ray into camera $j$'s coordinates, and $\pi_{K_j}$ denotes the pinhole projection with intrinsics $K_j$, mapping the ray to image coordinates $\mathbf q_j$. If multiple cameras cover the ray, we use the one whose optical axis is closest to it, and bilinearly sample its RGB image at $\mathbf q_j$.

\subsubsection{Generalization Evaluation Protocol}
To distinguish environmental robustness from cross-embodiment transfer, we vary lighting and robot embodiment separately and jointly while retaining the same twenty tasks. We train all models jointly on these tasks using 50 clean Aloha-AgileX demonstrations per task and a unified action and state representation. Evaluation crosses Aloha-AgileX and ARX-X5 with two lighting conditions: \emph{clean} retains default achromatic illumination, while \emph{randomized lighting} resamples point- and directional-light colors at each episode's start. The resulting four settings serve distinct purposes: clean Aloha establishes in-domain performance; randomized-lighting Aloha tests environmental robustness; clean ARX-X5 tests embodiment transfer; and randomized-lighting ARX-X5 tests both changes together. All generalization settings use the trained policies without additional training data or fine-tuning, so the comparisons measure transfer rather than adaptation to the target setting. For each task in each setting, we run 100 evaluation episodes and report the success rate; aggregate results are averaged over the twenty tasks.

\experimentcolorstrue
\section{EXPERIMENTS}
\label{sec:experiments}

\setcounter{dbltopnumber}{2}
\begin{figure*}[t]
  \centering
  \includegraphics[width=\textwidth,trim=0 0 0 -18bp]{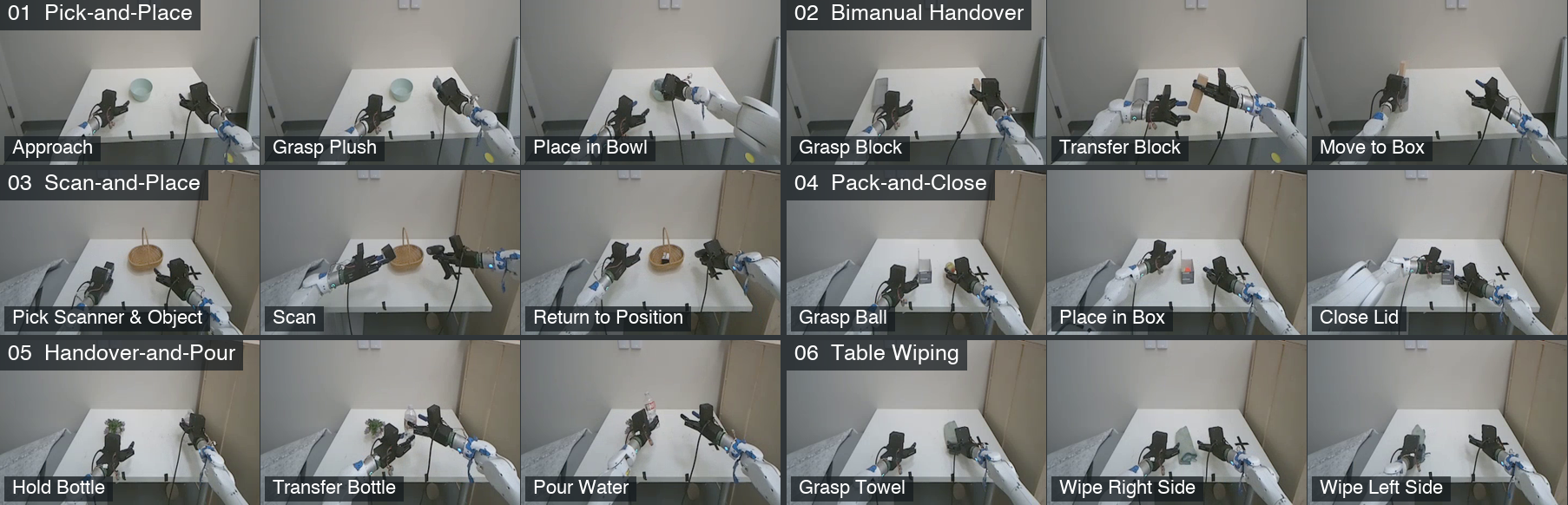}
  \caption{\textbf{Real-world task suite.} We show egocentric keyframes from six real-world tasks, including unimanual manipulation and bimanual coordination. These views are for visualization only and are not used as policy inputs.}
  \label{fig:real-world-tasks}
\end{figure*}

\begin{table*}[t]
  \centering
  \caption{\textbf{Real-world bimanual manipulation.} We report full-episode success rate (SR) and average progress score (Score), both in percent, over 20 evaluation episodes per task. Entries show SR / Score; Avg. is the unweighted mean over the six tasks.}
  \label{tab:real-world-results}
  \resizebox{0.94\textwidth}{!}{%
    \begin{tabular}{lccccccc}
      \toprule
      Method & Pick-and-Place & Handover & Scan-and-Place & Pack-and-Close & Handover-and-Pour & Table Wiping & Avg. \\
      \midrule
      \pifive      & 70.0 / 86.7 & 45.0 / 61.3 & 50.0 / 82.0 & 40.0 / 40.0 & 60.0 / 70.0 & \textbf{100.0} / \textbf{100.0} & 60.8 / 73.3 \\
      \rdtfm       & 70.0 / 76.7 & 5.0 / 35.0 & 5.0 / 42.0 & 30.0 / 47.5 & 45.0 / 63.0 & 85.0 / 92.5 & 40.0 / 59.4 \\
      \fastwam       & 50.0 / 81.7 & 40.0 / 71.3 & 45.0 / 67.0 & 30.0 / 42.5 & 25.0 / 45.0 & \textbf{100.0} / \textbf{100.0} & 48.3 / 67.9 \\
      \lingbot    & 60.0 / 86.7 & 5.0 / 35.0 & 55.0 / \textbf{91.0} & 40.0 / 55.0 & 25.0 / 67.0 & 75.0 / 95.8 & 43.3 / 71.8 \\
      \midrule
      \ours         & \textbf{95.0} / \textbf{98.3} & \textbf{80.0} / \textbf{93.8} & \textbf{75.0} / 87.0 & \textbf{60.0} / \textbf{80.0} & \textbf{100.0} / \textbf{100.0} & \textbf{100.0} / \textbf{100.0} & \textbf{85.0} / \textbf{93.2} \\
      \bottomrule
    \end{tabular}%
  }
\end{table*}

\begin{figure*}[t]
  \centering
  \begin{subfigure}[t]{0.55\textwidth}
    \centering
    \includegraphics[width=\linewidth]{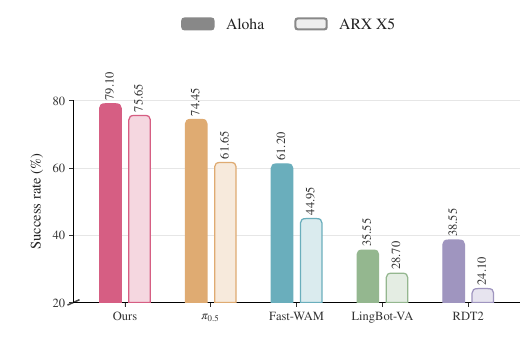}
    \caption{Average success across embodiments.}
    \label{fig:sim-results-mean}
  \end{subfigure}\hfill
  \begin{subfigure}[t]{0.43\textwidth}
    \centering
    \includegraphics[width=\linewidth]{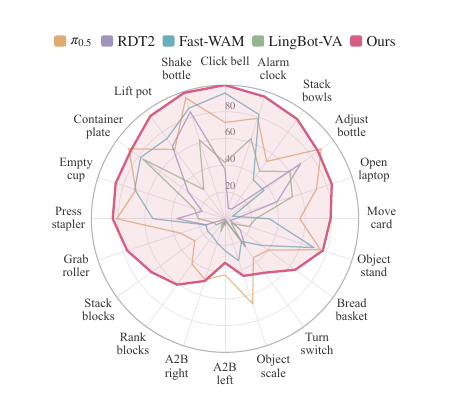}
    \caption{Cross-embodiment success on ARX X5.}
    \label{fig:sim-results-tasks}
  \end{subfigure}
\caption{\textbf{\simname evaluation across twenty tasks.} (a) Mean success rates on Aloha and ARX X5. (b) Per-task cross-embodiment success rates on ARX X5. \ours achieves the highest mean success rate on both embodiments and the smallest drop.}
  \label{fig:sim-results}
\end{figure*}

We evaluate the spatial WAM through closed-loop manipulation rather than standalone perceptual video metrics. Our experiments are designed to answer three questions:

\textbf{(Q1)} How does \ours compare with existing VLAs and WAMs on in-domain manipulation on \simname and in the real world? (Section~\ref{subsec:real-world-evaluation})

\textbf{(Q2)} Does \ours generalize better than baselines beyond the training conditions? (Section~\ref{subsec:generalization-evaluation})

\textbf{(Q3)} How do metric depth and two-stage UMI pretraining contribute to downstream manipulation performance? (Section~\ref{subsec:world-action-ablation})

\subsection{Experimental Setup}
\label{subsec:experimental-setup}

\subsubsection{Real-world tasks}
We construct a challenging real-world bimanual UMI evaluation on the Dexmate Vega1 mobile robot (Vega1), spanning precision-sensitive placement, coordinated handovers, tool use, long-horizon manipulation, liquid transfer, and sustained contact. As illustrated in \figref{fig:real-world-tasks}, its six tasks are \textbf{Pick-and-Place}, placing a plush toy into a bowl; \textbf{Bimanual Handover}, transferring a wooden block between grippers and into a box; \textbf{Scan-and-Place}, scanning and subsequently placing a product; \textbf{Pack-and-Close}, packing a baseball and closing its box; \textbf{Handover-and-Pour}, transferring a bottle between hands before watering flowers; and \textbf{Table Wiping}, wiping the table from right to left. For each task, we evaluate 20 episodes and report success rate (SR), the fraction of fully completed episodes, and progress score (Score), the average fraction of completed task stages. Both metrics are averaged over the six tasks.

\subsubsection{Baselines}
We compare \ours with a generalist VLA, \textbf{\pifive}~\cite{intelligence2025pi05visionlanguageactionmodelopenworld}, which learns manipulation skills from diverse robot data, and a UMI-specific VLA, \textbf{\rdtfm}~\cite{liu2026rdt2exploringscalinglimit}, which focuses on learning from UMI demonstrations. For WAMs, we choose \textbf{\fastwam}~\cite{yuan2026fastwam} and \textbf{\lingbot}~\cite{lingbot-va2026} as general-purpose WAM baselines. Baselines are initialized from their released base checkpoints, while \ours uses our spatial world-action pretrained checkpoint. On both \simname and real-world tasks, all methods are fine-tuned on the same demonstrations.

\subsubsection{Training protocol}
Across all evaluation setups, all policies use a 20-dimensional relative action $\mathbf{a} = [\Delta p_L,\, r_L^{6D},\, g_L,\, \Delta p_R,\, r_R^{6D},\, g_R]$, with chunk size 24, where $\Delta p \in \mathbb{R}^3$ denotes translation, $r^{6D} \in \mathbb{R}^6$~\cite{Zhou_2019_CVPR} denotes rotation, and $g \in [0, 0.1]$ denotes the absolute gripper opening width in meters.
For pretraining, we follow the two-stage procedure in Section~\ref{subsec:wam-training}, training each stage for 30,000 steps on the same \totalTime{}-hour RGB-D UMI dataset using 24 NVIDIA RTX PRO 6000 GPUs. For downstream tasks, we initialize \ours from its pretrained checkpoint and fine-tune it separately for real-world tasks and \simname, using 8 GPUs for each fine-tuning run and retaining the pretrained action representation.

\subsection{In-Domain Manipulation Performance on \simname and in Real-World Environments}
\label{subsec:real-world-evaluation}

To answer \textbf{Q1}, we compare in-domain manipulation performance in the real world and on \simname. Across six real-world tasks, \ours achieves 85.0\% average success, compared with 60.8\% for the strongest baseline, \pifive (\tabref{tab:real-world-results}). In Bimanual Handover, \pifive succeeds in 45.0\% of episodes, versus 80.0\% for our model. Transferring the wooden block requires \textbf{centimeter-level alignment} between the two UMI grippers. Competing policies exhibit gripper misalignment during transfer, highlighting the difficulty of estimating gripper--object distances from moving wrist RGB observations. In contrast, explicit metric depth provides our model with distance cues for aligning the receiving gripper with the held block.

The advantage extends to the twenty-task \simname benchmark. As shown in \figref{fig:sim-results-mean}, \ours achieves the highest in-domain average success rate of 79.10\%, outperforming the strongest VLA baseline, \pifive, at 74.45\% by 4.65 percentage points. The gap is even larger against competing WAMs: \fastwam reaches 61.20\% and \lingbot reaches 35.55\%, placing \ours 17.90 percentage points ahead of the strongest WAM baseline. Together, these results establish strong in-domain performance in simulation and the real world.

\subsection{Generalization Across Embodiments and Environments}
\label{subsec:generalization-evaluation}
\label{subsec:metric-geometry-ablation}

To answer \textbf{Q2}, we analyze the three generalization settings defined in Section~\ref{subsec:benchmark}. \figref{fig:sim-results-mean} reveals a clear separation that in-domain scores alone obscure. Although \ours and \pifive are separated by only 4.65 percentage points on Aloha, their gap widens to 14.00 percentage points after transfer to ARX X5. \ours achieves the highest cross-embodiment success rate with the smallest performance drop, showing that its advantage extends beyond strong performance on the training embodiment.

Cross-embodiment generalization requires preserving precise approach and contact behavior, not merely reaching the correct target. In a Click Alarmclock failure case, \fastwam approaches the clock but stops short of its top button, failing to establish the contact needed to trigger it. UMI pretraining equips \ours with manipulation priors that support fine-grained control beyond the training embodiment. World-only adaptation learns from UMI observations, while joint world-action pretraining connects these observations with end-effector actions, providing a foundation for transferring approach and contact skills to a different robot. The pretraining ablation in Section~\ref{subsec:world-action-ablation} further supports its contribution to cross-embodiment performance.

\begin{figure}[t]
  \centering
  \includegraphics[width=\columnwidth]{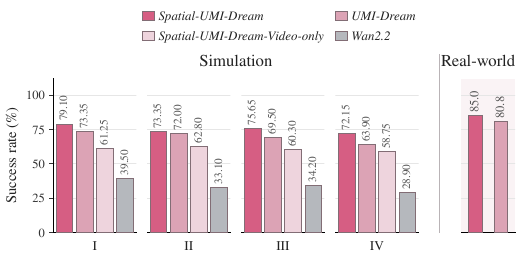}
  \caption{\textbf{Ablation of depth and UMI pretraining.} Simulation: I, Aloha clean; II, Aloha with randomized lighting; III, ARX X5 clean; IV, ARX X5 with randomized lighting. Bars report mean success over all twenty tasks. Real-world: mean success over six tasks for \textit{Spatial-UMI-Dream} (RGB-D) and \textit{UMI-Dream} (RGB).}
  \label{fig:main-analysis}
\end{figure}

\begin{table}[!b]
  \centering
  \caption{\textbf{Environment and joint-shift generalization on \simname.} Mean success rates (\%) over twenty tasks: Aloha under clean and randomized lighting, and ARX X5 under randomized lighting.}
  \label{tab:clean2rand-results}
  \small
  \begin{tabular*}{\columnwidth}{@{\extracolsep{\fill}}lccc@{}}
    \toprule
    Method & Aloha & Aloha & ARX X5 \\
           & Clean $\uparrow$ & Rand. Light $\uparrow$ & Rand. Light $\uparrow$ \\
    \midrule
    \pifive & 74.45 & 66.15 & 61.85 \\
    \rdt & 38.55 & 16.55 & 12.40 \\
    \fastwam & 61.20 & 18.85 & 16.90 \\
    \lingbot & 35.55 & 34.85 & 28.60 \\
    \midrule
    \ours & \textbf{79.10} & \textbf{73.35} & \textbf{72.15} \\
    \bottomrule
  \end{tabular*}
\end{table}

Under randomized lighting, Place Empty Cup exposes a different failure mode. The policy must grasp a cup from the table and place it on a coaster. In this case, \fastwam fails to localize the coaster accurately after illumination changes, leading to unsuccessful placement. This task requires maintaining spatial alignment with the placement target despite changes in its appearance. By modeling RGB together with metric depth, \ours uses explicit scene geometry alongside appearance cues to guide placement.

According to \tabref{tab:clean2rand-results}, \ours achieves 73.35\% success on Aloha with randomized lighting and 72.15\% under combined embodiment and environmental shifts, exceeding the strongest evaluated baseline by 7.20 and 10.30 percentage points, respectively.
Taken together, these results show that \ours sustains its in-domain advantage across embodiment and environmental shifts, with limited performance degradation.

\subsection{Ablation of Depth and Pretraining}
\label{subsec:world-action-ablation}

To answer \textbf{Q3}, we compare four variants using identical downstream demonstrations within each evaluation setup. \textit{UMI-Dream} and \textit{Spatial-UMI-Dream} share the same UMI demonstrations, two-stage training recipe, training budget, and $2\times2$ visual layout. They differ only in depth: the RGB-only variant duplicates its RGB views in place of depth panels to match the visual token count, while the full model uses aligned RGB-D observations. \textit{Spatial-UMI-Dream-Video-only} uses the same RGB-D data but only world-only adaptation, without action pretraining. The pretrained \textit{Wan2.2} model is directly fine-tuned on downstream RGB-D demonstrations without UMI pretraining.

\subsubsection{Depth enables more reliable control}
Metric depth improves success across all four simulation settings (\figref{fig:main-analysis}), by 5.75 percentage points in-domain and 8.25 under joint shifts. Stack Bowls Two requires moving two bowls to the table center and stacking them. Successful stacking depends on reaching each bowl at the correct distance and aligning it with the bowl underneath. The RGB-only policy misses grasps and misaligns the bowls, indicating difficulty in judging gripper--object distance and relative placement. Depth supplies explicit spatial cues for approaching grasp targets and aligning the bowls during placement. With demonstrations, training budget, and token count matched, these gains support \textbf{the value of geometric information rather than additional visual tokens}. On six real-world tasks, RGB-D raises mean success from 80.8\% to 85.0\% and improves mean progress.

\subsubsection{Two-stage pretraining improves downstream control}
World-only adaptation raises in-domain simulation success from 39.50\% (\textit{Wan2.2}) to 61.25\%, and joint world-action pretraining further raises it to 79.10\% (\figref{fig:main-analysis}). World-only adaptation exposes the pretrained backbone to moving UMI views and their metric geometry before downstream learning. Since \textit{Wan2.2} already provides a pretrained visual prior, this gain highlights the value of UMI-specific RGB-D experience even before action pretraining. Joint pretraining then couples these visual representations with end-effector actions, improving control beyond visual adaptation alone. The full model exceeds video-only pretraining by 15.35 and 13.40 percentage points under embodiment and joint shifts, respectively, and \textbf{performs best in all four settings}. These gains show that the benefit extends beyond the robot used for downstream training.

\experimentcolorsfalse
\section{CONCLUSION}
\label{sec:conclusion}

We introduced \ours, a 5B spatial WAM that jointly models RGB, fisheye-aligned metric depth, and actions through two-stage pretraining on \totalTime{} hours of UMI data. We also introduced \simname, a twenty-task benchmark for controlled evaluation of UMI policies across robot embodiments and environmental changes, separately and jointly. Experiments demonstrate strong in-domain performance and generalization, with 85.0\% mean success on six real-world tasks and the highest success rates across all benchmark settings. Matched RGB-only ablations establish the benefit of explicit metric geometry, while pretraining ablations show complementary gains from world-only adaptation and joint world-action learning. These gains extend to precise grasping and placement under moving wrist observations. Together, these results highlight the value of geometry-aware UMI pretraining for reliable manipulation and of controlled benchmarks for assessing policy generalization beyond training conditions.


\clearpage
\section*{APPENDIX}

\paragraph{Training recipe}
\ours is initialized from Wan2.2-TI2V-5B~\cite{wan2025wanopenadvancedlargescale}. The \totalTime{}-hour corpus combines two object-picking collections and a broader high-quality manipulation collection, sampled in proportion to their valid trajectory steps. World-only adaptation uses all source episodes, whereas joint world-action training uses only the action-supervised episodes retained after the filtering in Section~\ref{subsec:umi-pipeline}. In Eq.~\eqref{eq:depth-encoding}, $d_{\min}=0.01$\,m and $d_{\max}=1.5$\,m, and pixels without valid depth are encoded as black.

We train the full diffusion transformer and keep the text encoder, image encoder and video VAE frozen. All stages use AdamW with learning rate $10^{-5}$, $(\beta_1,\beta_2)=(0.95,0.999)$, $\epsilon=10^{-8}$ and weight decay $10^{-5}$, a cosine schedule with warmup over the first 5\% of updates, bfloat16, and DeepSpeed ZeRO-2. Joint world-action training and fine-tuning use $\lambda_a=1$. Table~\ref{tab:appendix-recipe} lists the stage-specific settings.

\begin{table}[h]
\centering
\caption{Stage-specific training settings.}
\label{tab:appendix-recipe}
\small
\begin{tabular*}{\columnwidth}{@{\extracolsep{\fill}}lrrr@{}}
\toprule
Stage & Updates & Batch & GPUs \\
\midrule
World-only adaptation & 30k & 192 & 24 \\
Joint world-action & 30k & 192 & 24 \\
Real-world fine-tuning & 22.5k & 64 & 8 \\
\simname fine-tuning & 10k & 128 & 8 \\
\bottomrule
\end{tabular*}
\end{table}

\paragraph{\simname setup}
Fisheye observations follow Eq.~\eqref{eq:fisheye-resampling}: the four co-centered source views are rendered at $320\times320$, with a $92^{\circ}$ forward view and $106^{\circ}$ auxiliary views, and resampled to a $320\times240$ KB4 image using the same left and right calibrations on both robots. For adaptive-grasp tasks, \simname computes the target gripper opening from the object and finger-pad collision geometry instead of using a fixed closure. It also replaces RoboTwin's per-arm cuRobo and MPLib planning with a PyRoKi formulation, solved with jaxls, that plans the twelve joints of both arms jointly, checks self- and environment collisions along the interpolated path, and time-parameterizes the result under velocity, acceleration and jerk limits.

For cross-embodiment evaluation, \simname fits the Aloha-AgileX and ARX-X5 models from the official RoboTwin release~\cite{chen2025robotwin} with the DAS/UMI gripper in place of their native grippers, adds calibrated wrist-fisheye RGB and depth cameras, and uses the planner above for both. The two embodiments therefore differ only in their URDFs, base transforms, home configurations and collision models. A single command $g\in[0,1]$ (0 closed, 1 open) drives the gripper's six mimic-coupled joints through a driving-joint angle of $0.925(1-g)$\,rad. The gripper is mounted at a $15^{\circ}$ pitch, and its TCP lies at $(0.0965,0,-0.043)$\,m in the gripper base frame.

For randomized-lighting evaluation, \simname reuses RoboTwin's light-color randomization: each point- and directional-light RGB color is drawn from $U[0,1)^3$ at episode start. RoboTwin's background, clutter, table-height and camera randomization stay disabled.

In every setting, the 100 evaluation episodes of a task use the first 100 seeds, counting up from 100000, for which the expert plan succeeds. An episode counts as a success if the task predicate holds at any physics step within the task's step limit (Table~\ref{tab:appendix-sim-tasks}).

\paragraph{Tasks}
Table~\ref{tab:appendix-sim-tasks} lists the twenty \simname tasks with step limits inherited from RoboTwin. They cover object placement (the seven Place tasks, Move Playingcard Away and Adjust Bottle), stacking and arrangement (Stack Blocks Two, Stack Bowls Two and Blocks Ranking RGB), articulated-object and button interaction (Open Laptop, Press Stapler, Turn Switch, Click Alarmclock and Click Bell), and dual-arm or dynamic manipulation (Grab Roller, Lift Pot and Shake Bottle). Each model is trained on 50 clean Aloha-AgileX demonstrations per task. For the six real-world tasks, Table~\ref{tab:appendix-real-data} lists the demonstrations and binary stages. The progress score is the fraction of stages completed, and an episode succeeds only if all stages are completed.

\begin{table}[t]
\centering
\caption{\simname tasks and step limits (policy actions).}
\label{tab:appendix-sim-tasks}
\footnotesize
\begin{tabular*}{\columnwidth}{@{\extracolsep{\fill}}>{\raggedright\arraybackslash}p{0.33\columnwidth}>{\raggedright\arraybackslash}p{0.50\columnwidth}r@{}}
\toprule
Task & Goal & Limit \\
\midrule
Adjust Bottle & Pick up the bottle upright with the appropriate arm & 400 \\
Blocks Ranking RGB & Arrange red, green and blue blocks left to right & 1200 \\
Click Alarmclock & Press the alarm clock's top button & 400 \\
Click Bell & Press the top of the bell & 400 \\
Grab Roller & Grasp and lift the roller with both arms & 400 \\
Lift Pot & Lift the pot with both arms & 400 \\
Move Playingcard Away & Move the playing card away from the table center & 400 \\
Open Laptop & Open the laptop lid & 1400 \\
Place A2B Left & Place object A to the left of object B & 400 \\
Place A2B Right & Place object A to the right of object B & 400 \\
Place Bread Basket & Put one or two breads into the basket & 700 \\
Place Container Plate & Place the container onto the plate & 400 \\
Place Empty Cup & Place the empty cup on the coaster & 500 \\
Place Object Scale & Put the object on the scale & 400 \\
Place Object Stand & Place the object on the stand & 400 \\
Press Stapler & Press down the stapler & 400 \\
Shake Bottle & Lift and shake the bottle & 700 \\
Stack Blocks Two & Stack the green block on the red block & 800 \\
Stack Bowls Two & Stack the two bowls & 900 \\
Turn Switch & Toggle the switch & 400 \\
\bottomrule
\end{tabular*}
\end{table}

\begin{table}[t]
\centering
\caption{Real-world demonstrations and scored stages.}
\label{tab:appendix-real-data}
\small
\setlength{\tabcolsep}{3pt}
\renewcommand{\arraystretch}{1.15}
\renewcommand{\tabularxcolumn}[1]{m{#1}}
\begin{tabularx}{\columnwidth}{@{}lrr>{\raggedright\arraybackslash\footnotesize}X@{}}
\toprule
Task & Demos & Frames & Scored stages \\
\midrule
Pick-and-Place & 20 & 3,350 & reach, grasp, place \\
Bimanual Handover & 20 & 11,565 & grasp, hand over, right-gripper release, place in box \\
Scan-and-Place & 40 & 18,278 & grasp product, grasp scanner, scan, put down scanner, place product \\
Pack-and-Close & 40 & 13,805 & grasp, place, close lid, press \\
Handover-and-Pour & 30 & 11,103 & reach, grasp, hand over, pour, release \\
Table Wiping & 30 & 10,931 & grasp, wipe, release, grasp, wipe, release \\
\midrule
Total & 180 & 69,032 & \\
\bottomrule
\end{tabularx}
\end{table}


\begin{thebibliography}{10}
\providecommand{\url}[1]{#1}
\csname url@samestyle\endcsname
\providecommand{\newblock}{\relax}
\providecommand{\bibinfo}[2]{#2}
\providecommand{\BIBentrySTDinterwordspacing}{\spaceskip=0pt\relax}
\providecommand{\BIBentryALTinterwordstretchfactor}{4}
\providecommand{\BIBentryALTinterwordspacing}{\spaceskip=\fontdimen2\font plus
\BIBentryALTinterwordstretchfactor\fontdimen3\font minus
  \fontdimen4\font\relax}
\providecommand{\BIBforeignlanguage}[2]{{%
\expandafter\ifx\csname l@#1\endcsname\relax
\typeout{** WARNING: IEEEtran.bst: No hyphenation pattern has been}%
\typeout{** loaded for the language `#1'. Using the pattern for}%
\typeout{** the default language instead.}%
\else
\language=\csname l@#1\endcsname
\fi
#2}}
\providecommand{\BIBdecl}{\relax}
\BIBdecl

\bibitem{chi2024universalmanipulationinterfaceinthewild}
C.~Chi \emph{et~al.}, ``Universal manipulation interface: In-the-wild robot
  teaching without in-the-wild robots,'' 2024, arXiv:2402.10329.

\bibitem{intelligence2025pi05visionlanguageactionmodelopenworld}
{Physical Intelligence} \emph{et~al.}, ``$\pi_{0.5}$: a vision-language-action
  model with open-world generalization,'' 2025, arXiv:2504.16054.

\bibitem{wei2026psi0}
S.~Wei \emph{et~al.}, ``{$\Psi_0$}: An open foundation model towards universal
  humanoid loco-manipulation,'' \emph{arXiv preprint arXiv:2603.12263}, 2026.

\bibitem{liu2026rdt2exploringscalinglimit}
S.~Liu \emph{et~al.}, ``{RDT2}: Exploring the scaling limit of {UMI} data
  towards zero-shot cross-embodiment generalization,'' 2026, arXiv:2602.03310.

\bibitem{kim24openvla}
M.~Kim \emph{et~al.}, ``{OpenVLA}: An open-source vision-language-action
  model,'' \emph{arXiv preprint arXiv:2406.09246}, 2024.

\bibitem{ye2026worldactionmodelszeroshot}
S.~Ye \emph{et~al.}, ``World action models are zero-shot policies,'' 2026,
  arXiv:2602.15922.

\bibitem{nvidia2026cosmos3omnimodalworld}
NVIDIA \emph{et~al.}, ``Cosmos 3: Omnimodal world models for physical {AI},''
  2026, arXiv:2606.02800.

\bibitem{dyna2026dyna2}
\BIBentryALTinterwordspacing
{Dyna Robotics}, ``Dyna-2: A 1-million-hour scaling law for world-action
  models,'' August 2026. [Online]. Available: \url{https://dyna.co/dyna-2}
\BIBentrySTDinterwordspacing

\bibitem{chen2025robotwin}
T.~Chen \emph{et~al.}, ``{RoboTwin 2.0}: A scalable data generator and
  benchmark with strong domain randomization for robust bimanual robotic
  manipulation,'' \emph{arXiv preprint arXiv:2506.18088}, 2025.

\bibitem{wu2026largerewardmodelsgeneralizable}
\BIBentryALTinterwordspacing
Y.~Wu \emph{et~al.}, ``Large reward models: Generalizable online robot reward
  generation with vision-language models,'' 2026. [Online]. Available:
  \url{https://arxiv.org/abs/2603.16065}
\BIBentrySTDinterwordspacing

\bibitem{intelligence2026pi07steerablegeneralistrobotic}
{Physical Intelligence} \emph{et~al.}, ``${\pi}_{0.7}$: a steerable generalist
  robotic foundation model with emergent capabilities,'' 2026,
  arXiv:2604.15483.

\bibitem{jia2026dreamplanefficientreinforcementfinetuning}
E.~Y.-T. Jia, W.~Yuan, T.~Shi, V.~Guizilini, J.~Mao, and Y.~Wang,
  ``{DreamPlan}: Efficient reinforcement fine-tuning of vision-language
  planners via video world models,'' 2026, arXiv:2603.16860.

\bibitem{zhang2026foreactsteeringvlaefficient}
Z.~Zhang \emph{et~al.}, ``{ForeAct}: Steering your {VLA} with efficient visual
  foresight planning,'' 2026, arXiv:2602.12322.

\bibitem{bi2025motusunifiedlatentaction}
H.~Bi \emph{et~al.}, ``Motus: A unified latent action world model,'' in
  \emph{Proceedings of the IEEE/CVF Conference on Computer Vision and Pattern
  Recognition}, 2026, pp. 35\,101--35\,113.

\bibitem{bevcalib}
W.~Yuan, J.~Li, J.~Yue, D.~Shah, K.~Karydis, and H.~Qiu, ``{BEVCALIB}:
  {LiDAR}-camera calibration via geometry-guided bird's-eye view
  representations,'' in \emph{9th Annual Conference on Robot Learning}, 2025.

\bibitem{Ze2024DP3}
Y.~Ze, G.~Zhang, K.~Zhang, C.~Hu, M.~Wang, and H.~Xu, ``{3D} diffusion policy:
  Generalizable visuomotor policy learning via simple {3D} representations,''
  in \emph{Proceedings of Robotics: Science and Systems (RSS)}, 2024.

\bibitem{3d_diffuser_actor}
\BIBentryALTinterwordspacing
T.-W. Ke, N.~Gkanatsios, and K.~Fragkiadaki, ``{3D} diffuser actor: Policy
  diffusion with {3D} scene representations,'' 2024. [Online]. Available:
  \url{https://arxiv.org/abs/2402.10885}
\BIBentrySTDinterwordspacing

\bibitem{nguyen2026iclrincontextimitationlearning}
\BIBentryALTinterwordspacing
T.~Nguyen, W.~Yuan, S.~Wei, H.~Li, D.~Seita, and Y.~Wang, ``{ICLR}: In-context
  imitation learning with visual reasoning,'' 2026. [Online]. Available:
  \url{https://arxiv.org/abs/2603.07530}
\BIBentrySTDinterwordspacing

\bibitem{zhen20243dvla}
H.~Zhen \emph{et~al.}, ``{3D-VLA}: A {3D} vision-language-action generative
  world model,'' 2024, arXiv:2403.09631.

\bibitem{qu2025spatialvla}
D.~Qu \emph{et~al.}, ``{SpatialVLA}: Exploring spatial representations for
  visual-language-action model,'' 2025, arXiv:2501.15830.

\bibitem{li2026spatialforcing}
\BIBentryALTinterwordspacing
F.~Li \emph{et~al.}, ``Spatial forcing: Implicit spatial representation
  alignment for vision-language-action model,'' 2025. [Online]. Available:
  \url{https://arxiv.org/abs/2510.12276}
\BIBentrySTDinterwordspacing

\bibitem{sun2026rocket}
G.~Sun \emph{et~al.}, ``{ROCKET}: Residual-oriented multi-layer alignment for
  spatially-aware vision-language-action models,'' \emph{arXiv preprint
  arXiv:2602.17951}, 2026.

\bibitem{zhao2026humanoideverydaycomprehensiverobotic}
\BIBentryALTinterwordspacing
Z.~Zhao \emph{et~al.}, ``Humanoid everyday: A comprehensive robotic dataset for
  open-world humanoid manipulation,'' 2026. [Online]. Available:
  \url{https://arxiv.org/abs/2510.08807}
\BIBentrySTDinterwordspacing

\bibitem{wei2026simple}
S.~Wei \emph{et~al.}, ``{SIMPLE}: Simulation-based policy learning and
  evaluation for humanoid loco-manipulation,'' \emph{arXiv preprint
  arXiv:2606.08278}, 2026.

\bibitem{generalist2026gen15}
{Generalist Team}, ``{GEN-1.5}: Embodied foundation models are one-shot
  learners,'' \emph{Generalist AI Blog}, 2026,
  https://generalistai.com/blog/gen-1.5.

\bibitem{zhang2026hy}
H.~Zhang \emph{et~al.}, ``{Hy-Embodied-0.5-VLA}: From vision-language-action
  models to a real-world robot learning stack,'' \emph{arXiv preprint
  arXiv:2606.14409}, 2026.

\bibitem{yang2026vista}
S.~Yang \emph{et~al.}, ``{VISTA}: Vision-grounded and physics-validated
  adaptation of {UMI} data for {VLA} training,'' \emph{arXiv preprint
  arXiv:2606.04708}, 2026.

\bibitem{wen2026fastfoundationstereo}
B.~Wen, S.~Dewan, and S.~Birchfield, ``{Fast-FoundationStereo}: Real-time
  zero-shot stereo matching,'' \emph{CVPR}, 2026.

\bibitem{ke2023repurposing}
B.~Ke \emph{et~al.}, ``Repurposing diffusion-based image generators for
  monocular depth estimation,'' in \emph{Proceedings of the IEEE/CVF Conference
  on Computer Vision and Pattern Recognition (CVPR)}, 2024.

\bibitem{wan2025wanopenadvancedlargescale}
{Team Wan} \emph{et~al.}, ``Wan: Open and advanced large-scale video generative
  models,'' 2025, arXiv:2503.20314.

\bibitem{lipman2023flowmatchinggenerativemodeling}
Y.~Lipman, R.~T.~Q. Chen, H.~Ben-Hamu, M.~Nickel, and M.~Le, ``Flow matching
  for generative modeling,'' 2023, arXiv:2210.02747.

\bibitem{tao2024maniskill3}
S.~Tao \emph{et~al.}, ``{ManiSkill3}: {GPU} parallelized robotics simulation
  and rendering for generalizable embodied {AI},'' \emph{arXiv preprint
  arXiv:2410.00425}, 2024.

\bibitem{yang2026realtimevlav2learningrun}
C.~Yang, Y.~Hu, Y.~Ma, Y.~Yang, J.~Tan, and H.~Fan, ``{Realtime-VLA V2}:
  Learning to run {VLAs} fast, smooth, and accurate,'' 2026, arXiv:2603.26360.

\bibitem{yuan2026fastwam}
T.~Yuan, Z.~Dong, Y.~Liu, and H.~Zhao, ``{Fast-WAM}: Do world action models
  need test-time future imagination?'' \emph{arXiv preprint arXiv:2603.16666},
  2026.

\bibitem{lingbot-va2026}
L.~Li \emph{et~al.}, ``Causal world modeling for robot control,'' \emph{arXiv
  preprint arXiv:2601.21998}, 2026.

\bibitem{Zhou_2019_CVPR}
Y.~Zhou, C.~Barnes, J.~Lu, J.~Yang, and H.~Li, ``On the continuity of rotation
  representations in neural networks,'' in \emph{Proceedings of the IEEE/CVF
  Conference on Computer Vision and Pattern Recognition (CVPR)}, June 2019.

\end{thebibliography}
\end{document}